\documentclass[11pt]{article}

\usepackage[margin=1in]{geometry}
\usepackage{amsmath, amssymb}
\usepackage{graphicx}
\usepackage{booktabs}
\usepackage{hyperref}
\usepackage{authblk}
\usepackage{natbib}
\usepackage{listings}
\usepackage{xcolor}
\usepackage{marvosym}

\title{\textbf{AnovaX: A Local, Multi-Agent Voice Assistant \\
with LLM Planning, Typed Executors, and Adaptive Recovery}}
\author[1]{Raunak B Sinha}
\affil[1]{BITS Pilani, India}
\date{\today}

\begin{document}

\maketitle

\begin{abstract}
Desktop voice assistants are still dominated by cloud pipelines that
ship raw audio off the machine and expose a fixed set of skills. We
describe AnovaX, a small local-first assistant that runs entirely on
the user's computer and treats the desktop itself as its action
surface. A single Python process wires together a wake-word gate, a
speech pipeline, an LLM planner (Gemini) that emits a JSON plan of
tool calls, a whitelist-and-denylist safety layer, a multi-agent
orchestrator that translates each plan into typed child agents on a
bounded thread pool, and an adaptive recovery loop that takes over
whenever a core step fails. Every tool corresponds to a specialized
agent class (\texttt{AppAgent}, \texttt{TypingAgent},
\texttt{BrowserAgent} and six others) with its own timeout, retry
policy, and shared-resource locks. A recursive \texttt{MetaAgent}
lets the planner delegate a sub-goal back to itself, capped at two
levels of nesting. The recovery loop uses a compact ReAct-style
prompt and hides Gemini's latency behind speculative execution of
read-only tools. A companion Flask server exposes a phone-friendly
remote over the local WiFi, mirrors every agent lifecycle event to
the phone in real time, and streams the laptop's screen back over
MJPEG so the user can watch remote commands land as they run. The
point of the project is less to compete with Siri or Alexa than to
show that a legible, few-thousand-line assistant is enough to open
apps, type into them, run searches, coordinate concurrent actions,
recover from single-step failures, and be driven entirely from a
phone in another room --- without the LLM ever touching the
keyboard.
\end{abstract}

\section{Introduction}
The interesting question about a desktop voice assistant, in 2025 or
2026, is no longer ``can a language model understand what I said.''
That has been settled for a while. The interesting question is
whether the assistant can \emph{act} on the machine in front of the
user, and whether the pieces that do the acting are simple enough
that a normal person can read them, fix them, and trust them.

Most of the assistants in widespread use --- Siri, Alexa, Google
Assistant, Copilot --- do not directly address that question. They
run in the cloud, they treat the desktop as an app store rather than
a canvas, and the user has no visibility into either the transcript
that leaves the microphone or the reasoning that produces the reply.
That is a reasonable engineering trade-off for a consumer product,
but it leaves a gap for people who want three specific properties at
once: (i) audio and personal facts stay on the machine unless the
user opts in; (ii) the assistant can open a real desktop application,
type into it, and press a hotkey, not just answer trivia; and (iii)
every tool call the LLM asks for is inspectable, whitelisted, and
rejectable before it runs.

We built AnovaX to fill exactly that gap. Its earlier revision was a
straight-line executor: the LLM produced a plan, a Python function
walked the plan step by step, and that was it. Simple to read, but
fragile in one specific way --- a stuck step blocked the rest of the
plan, and there was no way to parallelize independent actions or to
recover from a single failure. This paper describes the current
revision, which replaces the linear executor with a small multi-agent
orchestrator. Each tool in the plan schema is now backed by a typed
child agent class with its own timeout, retry policy, and set of
shared-resource locks. Plans still come from a single LLM call, but
they are dispatched, batched, and supervised by a Python parent that
never gives the model the keyboard.

The revision also adds a second, adaptive execution path. When the
static plan hits a core failure, an autonomous loop takes over: a
compact ReAct-style prompt asks Gemini for the next batch of
actions, safety-checks it, and hands it back to the same
orchestrator, iterating until the goal is met or a small budget is
exhausted. The loop hides Gemini's latency behind speculative
parallelism, pre-running read-only tools while the planner is still
deciding what to do next.

The design deliberately borrows the vocabulary of the multi-agent
literature --- child agents, orchestrator, lifecycle events, parallel
batches --- without adopting its scale. AnovaX is a small-scale
system of roughly 1{,}800 lines of Python and one HTML file, and
we care much more about legibility than about squeezing another point
of accuracy out of a benchmark.

\paragraph{Contributions.} This is a system description, not a
benchmark paper. In short:
\begin{itemize}
    \item A local, LLM-planned desktop voice agent whose executor
    is a small multi-agent orchestrator. Each of the ten tools in
    the plan schema maps to a typed child-agent class with its own
    TTL, retry policy, and shared-resource locks. The orchestrator
    can spawn as many children as the plan calls for; concurrency
    is capped at eight workers as a project-specific threshold.
    \item A two-stage safety filter --- prompt-level rules plus a
    Python whitelist + denylist --- that runs on every plan,
    including recursively generated sub-plans, before any child
    agent is spawned.
    \item An adaptive recovery layer built on batched ReAct with
    \emph{speculative parallelism}: when a core step fails, an
    autonomous loop takes over, and read-only tools from the
    remaining plan pre-run in the background while the planner
    decides the next batch.
    \item A three-layer memory and a lifecycle event bus that
    mirrors every agent state transition to the UI, an on-disk
    JSONL log, and the mobile remote --- which also streams the
    laptop's screen back to the phone via MJPEG.
\end{itemize}
The evaluation in Section~4 is qualitative. Section~5 documents
the current limitations and known failure modes.

\section{Related Work}

\paragraph{Commercial voice assistants.} Siri, Alexa, Google
Assistant and Cortana all pair a wake word with a cloud-hosted
intent parser and a fixed skill catalogue. Their scope has crept
toward LLM-backed responses \citep{gemini2023}, but the action
surface stays narrow and closed. The user cannot add a new skill in
twenty lines, and the audio is not local. AnovaX takes the opposite
side of both choices.

\paragraph{LLM agents that use tools.} There is a large recent
literature on wrapping an LLM in a loop that lets it call tools:
ReAct \citep{yao2023react} interleaves reasoning traces with tool
invocations; HuggingGPT \citep{shen2023hugginggpt} routes to
task-specific models; AutoGen \citep{wu2024autogen} composes
multiple LLM agents; Voyager \citep{wang2023voyager} shows a similar
plan-and-act pattern in Minecraft; and EvoAgent
\citep{yuan2024evoagent} generates specialized child agents
automatically via evolutionary operators. AnovaX borrows selectively
from this literature. The default path is not a ReAct loop --- the
LLM plans once, and a Python orchestrator (not the model) decides
which typed child agent handles which tool, which children can run
in parallel, and which ones should be killed after their TTL. Only
when a core step fails does the assistant fall back to a
ReAct-style loop (Section~3.5), and even there the loop is bounded
(six turns, twenty agents total) and every batch of proposed actions
goes through the same safety filter. The child-agent classes
themselves are hand-written and typed, not generated. This is a
deliberate choice: at desktop scale, the control code should be
short enough to audit, and the LLM should not be trusted to route
around its own mistakes.

\paragraph{Desktop and web UI automation.} On the web-agent side,
WebGPT \citep{nakano2021webgpt} and Mind2Web \citep{deng2023mind2web}
demonstrated LLM-driven browser control; on the phone side, Android
in the Wild \citep{rawles2023android} released a large dataset of
task demonstrations. These systems generally rely on DOM- or
accessibility-tree inputs to close the loop. AnovaX is sightless:
its typed executors fire keyboard and mouse events through
\texttt{pyautogui} and never inspect the screen. This is a weakness
(Section~5) but also the reason each executor stays under fifty
lines.

\paragraph{Local, privacy-first assistants.} Open-source projects
such as Mycroft, Rhasspy and Leon showed that a fully local voice
assistant is buildable, though most predate the LLM planning wave
and use hand-written intent parsers. AnovaX sits between those
projects and the newer agent literature: it keeps the local-first
stance but uses an LLM as the intent parser and planner, and a
small multi-agent runtime as the actuator.

\section{System Design}

\begin{figure}[h]
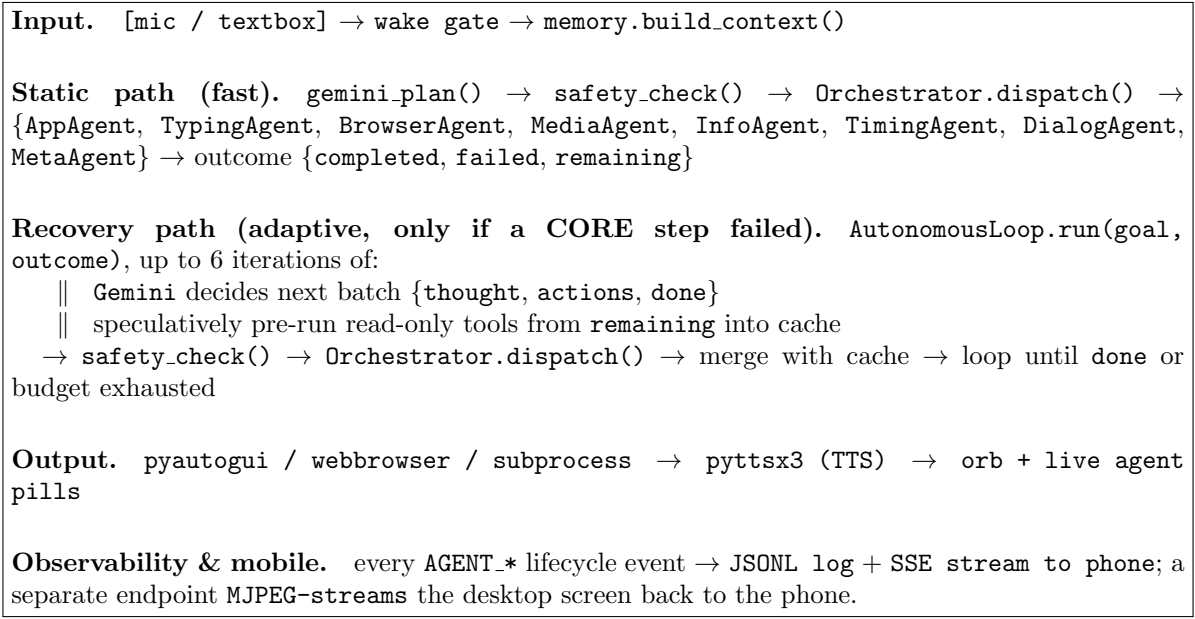

    \centering
    \fbox{\parbox{0.94\textwidth}{\small
    \textbf{Input.}\quad\texttt{[mic / textbox]} $\rightarrow$
    \texttt{wake gate} $\rightarrow$
    \texttt{memory.build\_context()}\\[3pt]

    \textbf{Static path (fast).}\quad
    \texttt{gemini\_plan()} $\rightarrow$
    \texttt{safety\_check()} $\rightarrow$
    \texttt{Orchestrator.dispatch()} $\rightarrow$
    \{\texttt{AppAgent}, \texttt{TypingAgent},
    \texttt{BrowserAgent}, \texttt{MediaAgent},
    \texttt{InfoAgent}, \texttt{TimingAgent},
    \texttt{DialogAgent}, \texttt{MetaAgent}\}
    $\rightarrow$ outcome \{\texttt{completed}, \texttt{failed},
    \texttt{remaining}\}\\[3pt]

    \textbf{Recovery path (adaptive, only if a CORE step failed).}\quad
    \texttt{AutonomousLoop.run(goal, outcome)}, up to 6 iterations of:\\
    \hspace*{4mm}$\;\;\parallel\;\;$ \texttt{Gemini} decides next batch
    \{\texttt{thought}, \texttt{actions}, \texttt{done}\}\\
    \hspace*{4mm}$\;\;\parallel\;\;$ speculatively pre-run read-only
    tools from \texttt{remaining} into cache\\
    \hspace*{4mm}$\rightarrow$ \texttt{safety\_check()} $\rightarrow$
    \texttt{Orchestrator.dispatch()} $\rightarrow$ merge with cache
    $\rightarrow$ loop until \texttt{done} or budget exhausted\\[3pt]

    \textbf{Output.}\quad
    \texttt{pyautogui / webbrowser / subprocess}
    $\rightarrow$ \texttt{pyttsx3 (TTS)}
    $\rightarrow$ \texttt{orb + live agent pills}\\[3pt]

    \textbf{Observability \& mobile.}\quad every \texttt{AGENT\_*}
    lifecycle event $\rightarrow$ \texttt{JSONL log}
    $+$ \texttt{SSE stream to phone};
    a separate endpoint \texttt{MJPEG-streams} the desktop screen
    back to the phone.
    }}
    \caption{End-to-end path a spoken request takes through AnovaX.
    Two execution paths share the same safety filter and the same
    orchestrator: a fast static path where the LLM plans once, and
    an adaptive recovery path that engages only when a core step
    fails and hides Gemini's latency behind speculative execution
    of read-only tools. The Gemini calls are the only steps that
    leave the machine, and they can be disabled --- a regex
    fallback covers the common commands.}
    \label{fig:arch}
\end{figure}

Figure~\ref{fig:arch} illustrates the overall architecture. Everything except
the Gemini call runs on the user's machine. Even the Gemini call is
optional --- if no API key is set, AnovaX falls back to a regex
intent parser that covers the most common commands.

\subsection{Wake-word gating and speech input}
The wake gate is intentionally minimal in its logic. A background thread runs
Google's free speech-recognition endpoint via the
\texttt{SpeechRecognition} library and only reacts to a phrase if it
contains the token \emph{anova} together with one of a small set of
trigger words (\emph{wake up, hey, hello, hi, yo}). While asleep,
all other utterances are dropped. This matters more than it looks:
a room full of ambient chatter cannot accidentally trigger an
action, and the model is never asked to plan for input the user did
not address to it.

Once awake, every recognized phrase is routed to a single
\texttt{on\_speech(text)} entry point. Text typed into the UI box
enters the same function, so the desktop and mobile inputs converge
before planning starts. This is also useful when the microphone is
absent (no PyAudio, VM without an audio device); AnovaX still runs,
just typed.

\subsection{The plan schema}
The LLM sees a system prompt that pins the reply to a fixed JSON
shape:

\begin{lstlisting}[language={}]
{
  "steps": [ {"tool": "<name>", "params": { ... }}, ... ],
  "final_response": "<short spoken reply>",
  "remember": { "<key>": "<value>" }
}
\end{lstlisting}

The tool set is intentionally short: nine concrete action tools
(\texttt{open\_app}, \texttt{type\_text}, \texttt{press\_keys},
\texttt{web\_search}, \texttt{youtube\_search}, \texttt{screenshot},
\texttt{get\_time}, \texttt{get\_date}, \texttt{wait},
\texttt{speak}) and one recursive-delegation tool
(\texttt{plan\_subtask}) that hands a natural-language sub-goal back
to the planner. Anything richer than that --- reading the current
window title, inspecting a file, running a shell command --- is not
in the schema. Adding a new tool requires editing three places
(the system prompt, the whitelist, and the executor mapping in
\texttt{agents.py}), and we wanted that cost to stay visible.

Tools split further into two runtime classes. \texttt{screenshot},
\texttt{get\_time}, \texttt{get\_date} and \texttt{wait} are
marked \emph{optional}: if one of them fails, the orchestrator
records the miss and moves on. Everything else is \emph{core}: a
failure halts the static plan and hands control to the adaptive
recovery loop described in Section~3.5. The split matters because
optional failures are usually harmless (a screenshot that couldn't
save, a clock read on a locked screen) and shouldn't disrupt the
rest of a compound request.

The prompt also asks the planner to insert a
\texttt{\{"tool":"wait","params":\{"seconds":1.5\}\}} between any
\texttt{open\_app} and a following \texttt{type\_text}. This is
purely a race-condition patch: on a cold machine, Notepad takes
noticeably longer to grab focus than \texttt{pyautogui.write()}
takes to fire, and without the wait the text is typed into
whichever window previously held focus. The multi-agent runtime has
narrowed this window with locks (Section~3.4), but the prompt-level
guard remains.

\subsection{Safety: two independent filters}
Two pieces stand between the LLM and the operating system.

The first is inside the system prompt: a short, explicit list of
actions the planner is instructed never to produce (delete, format, kill,
edit the registry, reveal secrets, and so on). This is a soft
filter, and we do not rely on it in isolation.

The second is a Python function, \texttt{safety\_check()}, that
runs on every plan the LLM returns --- including every sub-plan
generated by a \texttt{MetaAgent}. It enforces three properties:

\begin{enumerate}
    \item \textbf{Whitelist of tools.} Any step whose \texttt{tool}
    field is not in the eleven-item \texttt{ALLOWED\_TOOLS} set
    causes the whole plan to be rejected.
    \item \textbf{Denylist of keywords} on both the original user
    text and every string-valued parameter. The denylist covers
    roughly thirty patterns spanning file destruction
    (\texttt{rm -rf}, \texttt{format c:}), credential access
    (\texttt{password}, \texttt{keychain}), privilege elevation
    (\texttt{sudo}, \texttt{runas}), and OS-level state changes
    (\texttt{netsh}, \texttt{diskpart}, \texttt{regedit}). A single
    hit anywhere rejects the plan.
    \item \textbf{Bounded plan size.} At most eight steps per plan;
    at most five seconds per \texttt{wait}; at most two levels of
    \texttt{plan\_subtask} nesting; at most twenty child agents
    dispatched per user command in total.
\end{enumerate}

If a plan fails any check, no child agent is ever spawned and the
user hears a one-line refusal. Because the same filter re-runs on
every recursive sub-plan, a jailbreak that escapes the first-level
prompt still has to escape the whitelist a second time on the way
back in.

\subsection{Multi-agent orchestrator}
The heart of this revision is the orchestrator. When
\texttt{safety\_check()} passes, the plan is handed to
\texttt{Orchestrator.dispatch()}, which does four things in order:
persist any \texttt{remember} facts, split the step list into
sequential batches, spawn one typed child agent per step, and speak
the final reply once every batch has settled.

The orchestrator does not itself fix how many child agents a plan
can produce --- that number is determined by the LLM's plan,
subject only to the safety filter's per-plan step cap and the
per-command agent budget. What \emph{is} bounded is concurrency: in
the current build we run at most eight child agents at any moment,
because eight was the size at which we could still trace the
interactions between them by eye. A plan with three tool calls
spawns three agents; a plan with twenty spawns as many as needed
but runs at most eight of them in flight. Both numbers are
project-specific thresholds, not architectural limits.

Each tool in the schema has a corresponding \texttt{ChildAgent}
subclass. Table~\ref{tab:agents} lists them. Every subclass
inherits a common lifecycle (\texttt{CREATED} $\rightarrow$
\texttt{RUNNING} $\rightarrow$ \texttt{DONE}/\texttt{FAILED}/\texttt{KILLED}),
a per-class default TTL, a per-class retry budget, and a set of
named locks that must be held while \texttt{do\_work()} runs.

\begin{table}[h]
    \centering
    \small
    \begin{tabular}{lllccl}
        \toprule
        Class & Tool(s) & TTL (s) & Retries & Locks & Purpose \\
        \midrule
        \texttt{AppAgent}     & \texttt{open\_app}      & 10 & 1 & --- & launch desktop app \\
        \texttt{TypingAgent}  & \texttt{type\_text}, \texttt{press\_keys} & 30 & 0 & \texttt{kb\_mouse} & keyboard input \\
        \texttt{BrowserAgent} & \texttt{web\_search}, \texttt{youtube\_search} & 5 & 2 & --- & open URL \\
        \texttt{MediaAgent}   & \texttt{screenshot}     & 5  & 0 & --- & capture screen \\
        \texttt{InfoAgent}    & \texttt{get\_time}, \texttt{get\_date} & 2 & 0 & --- & read-only clock \\
        \texttt{TimingAgent}  & \texttt{wait}           & $\propto$secs & 0 & --- & sleep \\
        \texttt{DialogAgent}  & \texttt{speak}          & 30 & 0 & \texttt{tts} & TTS \\
        \texttt{MetaAgent}    & \texttt{plan\_subtask}  & 45 & 0 & --- & recursive planning \\
        \bottomrule
    \end{tabular}
    \caption{Typed child agents and their runtime properties. Every
    agent's TTL is additionally capped by a global 60-second ceiling.}
    \label{tab:agents}
\end{table}

\paragraph{Batching.} Read-only tools (\texttt{get\_time},
\texttt{get\_date}, \texttt{screenshot}) are marked
parallel-safe. The batcher walks the plan and groups consecutive
parallel-safe steps into a single batch that is fanned out on the
thread pool; every other step is its own batch. In practice, most
plans are one step per batch, but when a request asks for e.g.
``take a screenshot and tell me the time,'' the two calls issue in
parallel.

\paragraph{Locks.} Two named locks arbitrate the shared resources
that cannot be parallelized. The \texttt{kb\_mouse} lock is held by
any agent that will type or fire a hotkey, so a
\texttt{TypingAgent} never races with another
\texttt{TypingAgent}. The \texttt{tts} lock is held by
\texttt{DialogAgent}, so two concurrent \texttt{speak} calls
serialize instead of talking over each other. Locks are acquired
in a stable alphabetical order to eliminate the possibility of
deadlock between agents that need both.

\paragraph{TTL and retries.} A janitor thread runs on a 500 ms
tick, iterates over the currently running agents, and transitions
any agent whose elapsed time exceeds its TTL into the \texttt{KILLED}
state. Its result is logged and the failure surfaces on the UI and
mobile. Failures inside \texttt{do\_work()} are retried up to the
class's retry budget with exponential backoff
($0.5 \cdot 2^{n-1}$ seconds before attempt $n$). The retry budget
is chosen per class: browsers get two retries because DNS blips are
transient; typing gets zero retries because retrying a keypress
that half-succeeded is worse than failing.

\paragraph{Recursive planning.} A \texttt{MetaAgent} is a
\texttt{ChildAgent} that, in its \texttt{do\_work()}, re-invokes
\texttt{gemini\_plan()} on a sub-goal, re-runs
\texttt{safety\_check()} on the result, and hands the sub-plan
back to the same orchestrator with a bumped depth counter. Recursion
is hard-capped at two levels. This lets the planner say ``open my
inbox and type a two-paragraph reply'' as a single
\texttt{plan\_subtask} without exceeding the eight-step budget of
the top-level plan, while still going through the safety filter
twice on the way in.

\paragraph{Dispatch outcome.} Every call to
\texttt{Orchestrator.dispatch()} returns a small outcome dict:
\texttt{completed} (steps that ran, with their results and
durations), \texttt{failed} (the first core-step failure, if any,
plus its error), and \texttt{remaining} (the steps queued behind
the failure that were never attempted). If \texttt{failed} is
\texttt{None} the assistant speaks the final response and stops.
If \texttt{failed} is populated, control is handed to the adaptive
recovery loop.

\subsection{Adaptive recovery: the autonomous loop}
The static plan is the fast path, and it handles the vast majority
of real requests. But an LLM planner is not a compiler: it will
occasionally emit a step whose assumption is wrong --- an
application that isn't installed, a browser tab that lands on a
consent screen, a hotkey that a modal dialog steals. The
recovery layer is designed to notice, adapt, and finish the goal
without another round-trip to the user.

When \texttt{Orchestrator.dispatch()} returns with
\texttt{failed} populated on a core step, the \texttt{execute\_plan}
entry point invokes \texttt{AutonomousLoop.run()} with the original
user goal and the dispatch outcome. The loop uses a second, more
compact system prompt that asks Gemini for a batched
ReAct-style reply:

\begin{lstlisting}[language={}]
{
  "thought":  "<one-sentence plan>",
  "actions":  [ {"tool":"...", "params":{...}}, ... ],
  "done":     true | false,
  "summary":  "<final reply, if done>"
}
\end{lstlisting}

Each turn's \texttt{actions} run through the same
\texttt{safety\_check()} the static planner uses, and are then
dispatched back to the orchestrator as a fresh mini-plan.
\texttt{thought} is compacted into the history that the next turn
sees. The loop terminates when Gemini sets \texttt{done: true}, or
after a hard cap of six turns, or after a total budget of twenty
child agents across the whole recovery is exhausted.

\paragraph{Speculative parallelism.} A round-trip to Gemini costs
roughly one second on our machine (Section~4). Rather than let the
executor idle in that gap, the autoloop spawns a small speculation
pass in parallel with the planner call. The speculator scans the
\texttt{remaining} list from the failed static plan, picks any
step whose tool is read-only (\texttt{get\_time}, \texttt{get\_date},
\texttt{screenshot}), and runs it inline into a small cache keyed
by the step's normalized \texttt{(tool, params)} JSON. When
Gemini's decision comes back, any action whose key already sits in
the cache is served from there instead of being dispatched again.
In practice this hides the entire clock-read or screenshot latency
behind the planning turn --- the log line
``\texttt{AUTO~~\Lightning~2/3 action(s) served from speculation
cache}'' is the visible payoff.

\paragraph{Boundaries.} The recovery loop reuses the same safety
filter, the same orchestrator, and the same event bus as the
static path, so nothing new leaks out through the sides. What
changes is only where the plan comes from --- one prompt for the
opening move, a slightly different prompt for every recovery turn.
The recursive \texttt{MetaAgent} and the autoloop are the two
places where the LLM writes plans; both go through
\texttt{safety\_check()} on every entry.

\subsection{Memory in three layers}
The memory module is the most compact component with non-trivial behaviour. It keeps
three things:

\begin{itemize}
    \item A \emph{rolling history} of the last twenty conversational
    turns, kept in RAM only.
    \item A \emph{session state} recording which apps have been
    opened in the last five minutes, so ``type a paragraph in it''
    can resolve ``it'' without re-opening.
    \item A \emph{persistent facts} dictionary, saved to
    \texttt{memory.json}, that the planner is invited to write to
    via the \texttt{remember} field of its reply.
\end{itemize}

Only the third layer touches disk. The prompt is asked to keep the
keys short and lowercase (\emph{name}, \emph{role},
\emph{favorite\_editor}), and to leave out anything ephemeral or
sensitive. Before every Gemini call, the three layers are folded
into a plain-text CONTEXT block appended to the system prompt.
Session state is also read by \texttt{TypingAgent}: if the most
recently opened app was launched within the last three seconds,
the agent sleeps just long enough to let focus settle before firing
its first keypress. This is where the old open-then-type race
condition finally goes away.

\subsection{Observability and the mobile companion}
Every child agent publishes lifecycle events (\texttt{AGENT\_CREATED},
\texttt{AGENT\_STARTED}, \texttt{AGENT\_DONE}, \texttt{AGENT\_FAILED},
\texttt{AGENT\_KILLED}) through a small pub/sub bus. Three
subscribers listen: the desktop UI, which renders each active agent
as a colored pill in a live strip below the log; an on-disk JSONL
appender (\texttt{agent\_log.jsonl}); and the mobile server, which
forwards events to any connected phone via Server-Sent Events.

The phone remote is one Flask app on a background thread with four
endpoints (\texttt{/}, \texttt{/command}, \texttt{/wake},
\texttt{/sleep}) plus an \texttt{/events} stream. Auth is a shared
PIN set in the environment (\texttt{ANOVA\_MOBILE\_PIN}); the phone
sends it with every request and as a query parameter on the SSE
URL. The mobile page uses the browser's Web Speech API, which means
voice-to-text runs on the phone itself, not on the desktop.

We chose SSE over WebSockets because it survives most home routers,
needs no client library, and is one-directional, which matches our
threat model (the desktop is the trusted party). The important
addition in this revision is that the phone now watches the same
agent lifecycle events as the desktop, so when a \texttt{TypingAgent}
takes an unusual amount of time to acquire the \texttt{kb\_mouse}
lock, both surfaces show it at the same time.

\paragraph{MJPEG screen stream.} A separate endpoint on the same
Flask server offers a live view of the laptop's screen back to the
phone, encoded as a \emph{motion-JPEG} multipart HTTP response.
Every 150--300\,ms the desktop grabs the current screen with
\texttt{pyautogui.screenshot()}, downsizes it, JPEG-encodes it, and
appends the frame as a boundary-separated part. The phone's
browser renders the stream inside a plain \texttt{<img>} tag ---
which handles \texttt{multipart/x-mixed-replace} responses natively
--- so no player, no WebRTC, no polling loop is required on the
client side. The stream is served under the same PIN-authenticated
route as the command endpoints, and it can be paused by the user
to save bandwidth. In practice this means a user standing in
another room can speak a command into their phone, watch the
laptop's screen update in near real-time as the typed child agents
execute, and confirm the outcome without ever touching the
keyboard.

\section{Evaluation}

We treat this as a system paper and evaluate qualitatively.

\subsection{Setup}
All measurements were made on a single laptop: Windows 11 Pro,
Intel-class CPU, 16 GB RAM, on a residential broadband connection.
Voice input was captured with a built-in microphone at default
recognizer sensitivity. Gemini calls used
\texttt{gemini-flash-latest} through the public API. The offline
comparison uses the regex fallback with the Gemini path disabled.

\subsection{Capability coverage}
Table~\ref{tab:coverage} summarizes the classes of request AnovaX
handles. Coverage was checked by hand by running each row's example
phrase five times and marking it \emph{works} if the intended action
completed on at least four runs.

\begin{table}[h]
    \centering
    \begin{tabular}{lll}
        \toprule
        Category & Example phrase & Result \\
        \midrule
        Launch a desktop app & \emph{open notepad} & works \\
        Launch and type & \emph{open notepad and write hi} & works \\
        Launch a browser page & \emph{open youtube} & works \\
        Type into focused window & \emph{type i love coding} & works \\
        Web search & \emph{search for biryani recipes} & works \\
        YouTube search & \emph{play lofi beats on youtube} & works \\
        Time / date & \emph{what time is it} & works \\
        Screenshot to home & \emph{take a screenshot} & works \\
        Small talk / jokes & \emph{tell me a joke} & works \\
        Parallel read-only & \emph{screenshot and tell me the time} & works \\
        Recursive sub-plan & \emph{open notepad and draft me a short email} & works \\
        Multi-step compound & \emph{open calc, type 21+21, press enter} & works \\
        Refusal cases & \emph{delete my downloads folder} & correctly refused \\
        \bottomrule
    \end{tabular}
    \caption{Behaviour of AnovaX on a hand-checked set of request
    classes. The last three rows are new in this revision: they
    exercise the parallel batcher, the recursive
    \texttt{plan\_subtask}, and the \texttt{kb\_mouse} lock
    respectively.}
    \label{tab:coverage}
\end{table}

\subsection{Latency characteristics}
End-to-end latency on our machine is dominated by three components:
speech recognition (variable, network bound), the Gemini planning
call, and the offline TTS render. The orchestrator itself adds
almost nothing --- dispatching to the thread pool and enforcing
locks is measured in single-digit milliseconds. Rough ranges we
observed while watching the log:

\begin{table}[h]
    \centering
    \begin{tabular}{lc}
        \toprule
        Stage & Approx.\ time (s) \\
        \midrule
        Google speech recognition (short phrase) & 0.6--1.4 \\
        Gemini plan generation & 0.7--2.0 \\
        Safety check + orchestrator dispatch & $<$0.05 \\
        Child-agent lock acquisition (uncontended) & $<$0.01 \\
        \texttt{pyttsx3} render + speak (10-word reply) & 1.2--2.0 \\
        \bottomrule
    \end{tabular}
    \caption{Indicative per-stage latency. Numbers are ranges from
    a session log; they are not benchmark results and vary with
    network.}
    \label{tab:latency}
\end{table}

\subsection{A worked example}
Consider the phrase \emph{take a screenshot and tell me the time}.
This is the case where the batcher earns its keep, because both
tools are parallel-safe. The transcript from the log looks like
this (trimmed):

\begin{lstlisting}[language={}]
INPUT   awake=True text='take a screenshot and tell me the time'
GEMINI  response: {"steps":[
   {"tool":"screenshot","params":{}},
   {"tool":"get_time","params":{}}
 ], "final_response":"Done.", "remember":{}}
SAFETY  passed (2 step(s))
ORCH    dispatching 2 step(s) in 1 batch(es)
AGENT   MediaAgent-3a1c9d  RUNNING task=screenshot
AGENT   InfoAgent-b7e2f4   RUNNING task=get_time
AGENT   InfoAgent-b7e2f4   DONE (0.00s) {"time": "1:47 PM"}
AGENT   MediaAgent-3a1c9d  DONE (0.08s) {"path": "~/anova_screenshot_1751728077.png"}
SPEAK   "It's 1:47 PM."
\end{lstlisting}

Both agents entered \texttt{RUNNING} in the same batch, and the
final reply was overridden by the \texttt{InfoAgent}'s clock value
even though the plan's \texttt{final\_response} was the generic
``Done.''

\subsection{Ablations}
We tested four configurations by turning components off:

\begin{itemize}
    \item \textbf{Gemini disabled.} With no API key, the regex
    fallback handled 8 of the 13 rows in Table~\ref{tab:coverage}.
    It fails on the recursive sub-plan row, on paraphrased launches
    (e.g. \emph{fire up chrome}), on parallel batching, and on
    jokes it has not seen. The safety layer is unaffected because
    the fallback only knows how to call the same tools.
    \item \textbf{Memory disabled.} With the persistent facts
    layer cleared and the session state reset, ``it'' and ``that
    one'' references broke immediately: \emph{write a paragraph
    in it} sent the LLM back to \texttt{open\_app} with a guessed
    name, usually Notepad, sometimes Word.
    \item \textbf{Orchestrator locks disabled.} We disabled the
    \texttt{kb\_mouse} lock as a stress test and re-ran the
    multi-step compound row. On roughly one in five runs, a
    \texttt{press\_keys} agent fired its hotkey before the
    preceding \texttt{type\_text} agent had finished writing.
    \item \textbf{Autoloop disabled.} With the recovery loop
    turned off, a plan that included the wrong Chrome name
    (\emph{google chrome} on a machine where the launcher expects
    \emph{chrome}) failed the static \texttt{AppAgent} and
    surfaced ``I couldn't open google chrome.'' With the loop
    re-enabled, Gemini's next turn retried with
    \texttt{web\_search} instead and completed the goal in one
    additional round-trip. The speculative pre-run of
    \texttt{get\_time} shaved a further $\sim$0.7\,s off the
    perceived recovery latency on that request.
\end{itemize}

None of the ablations changed the safety behaviour --- refusals
continued to fire on the denylisted phrases in every mode. That is
the property we cared most about.

\section{Discussion and Limitations}

AnovaX is used daily by the author, but it exhibits several
limitations that a thorough treatment should not omit.

\paragraph{The executor is still blind.} The move to typed
per-tool agents did not add screen-content perception. When a step
goes wrong --- an app that didn't open, a modal dialog that stole
focus, a browser tab that landed on a redirect --- the offending
agent may complete its \texttt{do\_work()} without an exception
because it does not verify whether the intended effect occurred. The
rest of the plan then runs anyway, sometimes into the wrong window.
The lifecycle bus makes this failure \emph{visible}, which is a
real improvement over the previous revision, but it does not make
it \emph{recoverable} without a proper observe-act loop.

\paragraph{\texttt{pyautogui} is still fragile on Windows.} The
\texttt{kb\_mouse} lock removes intra-plan races between agents,
but it does not remove races between AnovaX and whatever the OS
happens to do next. On a cold boot, or when Windows Defender
scans a newly launched executable, the post-launch settle still
misses. Polling for the target window title before typing is the
principled fix; we have not implemented it yet.

\paragraph{Cloud dependencies where we could stay local.} Speech
recognition uses Google's endpoint and the planner uses Gemini.
Both are network dependencies, and both send text off the machine.
A private-first release should swap in Whisper (local) and a small
local LLM for planning. We kept the current pair because they
made the code short and the paper easy to write; they were not a
principled choice.

\paragraph{The multi-agent design is deliberately small in scale.}
Recursive planning is capped at two levels, the pool at eight
workers, and the per-command agent budget at twenty. These
numbers are not scientific --- they are the sizes at which we
could keep debugging the interactions in our head. A bigger
system would benefit from tracing (OpenTelemetry, say) and from
a fair-share scheduler across users; we have neither.

\paragraph{The recovery loop can chase a bad idea for six turns.}
The autoloop's budget of six iterations is short enough to fail
fast in practice, but it is not smart enough to notice when its
own \texttt{thought} keeps re-proposing minor variants of the same
broken step. We rely on the ``do not retry the exact same failed
step with the exact same params'' prompt rule, which is a soft
constraint. A proper trajectory-level check would be better; we
haven't written one.

\paragraph{The MJPEG stream leaks screen content.}
Any authenticated phone can see anything on the desktop screen.
This is exactly what the feature is for, but it means the PIN
becomes a much more valuable secret. A production version would
add a per-session token, a rate limit, and an idle timeout on the
stream; we currently have none of those.

\paragraph{The mobile remote is minimal.} PIN over HTTP is fine
on a trusted home network and inappropriate anywhere else. There
is no transport encryption, no rate limit, and no rotation. If
someone runs this on a laptop connected to a coffee-shop network
with the PIN set, the resulting exposure would be significant.

\paragraph{The safety filter is a blacklist.} Blacklists are the
weaker half of the pair with a whitelist, and we know it. The
tool whitelist is the load-bearing piece; the keyword denylist is
belt-and-suspenders. New attack vectors that stay inside the
allowed tools (say, opening a browser and navigating to a
phishing page) are not addressed at all. The recursive
\texttt{plan\_subtask} increases the attack surface by exactly
one round-trip through the LLM per level of nesting, which is why
the depth cap is two, not five.

\section{Conclusion}
AnovaX is a working demonstration that a locally-executing desktop
assistant can be built at modest scale by placing the LLM in a
planner role and keeping a small, whitelisted multi-agent runtime
between the plan and the operating system. The typed child agents,
the two-stage safety filter, the three-layer memory, the
recursive \texttt{MetaAgent}, the autonomous recovery loop with
speculative parallelism, the observability bus, and the MJPEG
screen stream to the phone are the pieces that make the resulting
system practical in daily use. None of the individual pieces
are novel. What we hope is copyable is the ratio: a single LLM
planning call as the fast path, a bounded ReAct loop as the slow
path, a handful of small typed executors, a conventional Python parent
process that handles concurrency, and a hard-to-jailbreak safety
filter that runs on every plan the model produces --- static,
recursive, or recovery.

The next step we care most about is a local model swap ---
Whisper for speech and a compact instruction-tuned local LLM for
planning --- so that an end user can run AnovaX on an offline
laptop, which is the configuration privacy-conscious users typically
require.

\paragraph{Remote voice control from the phone.} Because the mobile
companion (Section~3.6) exposes the same wake, sleep, and command
endpoints as the desktop, a user can leave the laptop in one room
and drive it entirely by voice from their phone anywhere on the
same WiFi network. Speech-to-text runs in the phone's own browser
via the Web Speech API, the transcript is posted to the desktop
over a PIN-authenticated HTTP endpoint, the orchestrator runs the
same typed child agents it would run for a local command, and every
lifecycle event streams back to the phone over Server-Sent Events
--- so the phone shows the same live agent pills as the desktop
does. In practice this means the laptop can sit on a desk while
its owner walks around the house and, from the phone, says
``AnovaX, open Chrome and search for train tickets to Delhi,'' and
watches the plan execute remotely. No cloud service is involved;
the phone talks to the laptop directly.

\bibliographystyle{plainnat}
\bibliography{references}

\end{document}